\documentclass[11pt,a4paper]{article}
\usepackage[utf8]{inputenc}
\usepackage{coling08}
\usepackage{times}
\usepackage{latexsym}
\usepackage{tree-dvips}
\usepackage{vaucanson-g}
\usepackage{url}
\usepackage{graphicx}
\newcommand{\dotted}[0]{\makedash{2pt}}

\title{TuLiPA: Towards a Multi-Formalism Parsing Environment for\\
  Grammar Engineering}

\author{Laura Kallmeyer \\
  SFB 441 \\
  Universit\"at T\"ubingen \\
  D-72074, T\"ubingen, Germany \\
  {\small \tt lk@sfs.uni-tuebingen.de~~~~} \\[2ex]
  {\bf Yannick Parmentier} \\
  CNRS - LORIA \\
  Nancy Université \\
  F-54506, Vand{\oe}uvre, France \\
  {\small \tt parmenti@loria.fr} \And
  Timm Lichte \\
  SFB 441 \\
  Universit\"at T\"ubingen \\
  D-72074, T\"ubingen, Germany \\
  {\small \tt ~~~~timm.lichte@uni-tuebingen.de~~~~} \\[2ex]
  {\bf Johannes Dellert} \\
  SFB 441 - SfS \\
  Universit\"at T\"ubingen \\
  D-72074, T\"ubingen, Germany \\
  {\small \tt ~~~~~~~~~~~~~~~~~~~~~~~~\{jdellert,kevang\}@sfs.uni-tuebingen.de} \And
  Wolfgang Maier \\
  SFB 441 \\
  Universit\"at T\"ubingen \\
  D-72074, T\"ubingen, Germany \\
  {\small \tt ~~~~wo.maier@uni-tuebingen.de} \\[2ex]
  {\bf Kilian Evang} \\
  SFB 441 - SfS \\
  Universit\"at T\"ubingen \\
  D-72074, T\"ubingen, Germany \\
  {\small \tt ~~~~}
}
\date{}

\begin{document}

\maketitle

\begin{abstract}
  In this paper, we present an open-source parsing environment
  (T\"ubingen Linguistic Parsing Architecture, TuLiPA) which uses Range
  Concatenation Grammar (RCG) as a pivot formalism, thus opening the
  way to the parsing of several mildly context-sensitive formalisms. This
  environment currently supports tree-based grammars (namely
  Tree-Adjoining Grammars (TAG) and Multi-Component Tree-Adjoining
  Grammars with Tree Tuples (TT-MCTAG)) and allows computation not
  only of syntactic structures, but also of the corresponding semantic
  representations. It is used for the development of a tree-based
  grammar for German.
\end{abstract}

\section{Introduction}
Grammars and lexicons represent important linguistic resources for
many NLP applications, among which one may cite dialog systems,
automatic summarization or machine translation. Developing such
resources is known to be a complex task that needs useful tools such
as parsers and generators \cite{erbach92tools}.

Furthermore, there is a lack of a common framework allowing for
multi-formalism grammar engineering. Thus, many formalisms have been
proposed to model natural language, each coming with specific
implementations. Having a common framework would facilitate the
comparison between formalisms (e.g., in terms of parsing
complexity in practice), and would allow for a better sharing of
resources (e.g., having a common lexicon, from which different
features would be extracted depending on the target formalism).

In this context, we present a parsing environment relying on a general
architecture that can be used for parsing with mildly
context-sensitive (MCS) formalisms\footnote{A formalism is said to be 
  mildly context sensitive (MCS) iff (i)~it generates limited
  cross-serial dependencies, (ii)~it is polynomially parsable, and 
(iii)~the string languages generated by the formalism have the
constant growth property (e.g., $\{a^{2^n} | n \geq 0\}$ does not
have this property). Examples of MCS formalisms include Tree-Adjoining
Grammars, Combinatory Categorial Grammars and Linear Indexed
Grammars.} \cite{Joshi:87}. Its underlying idea is to use Range
Concatenation Grammar (RCG) as a pivot formalism, for RCG has been
shown to strictly include MCS languages while being parsable in
polynomial time \cite{Boullier:00}. 

Currently, this architecture supports tree-based grammars
(Tree-Adjoining Grammars and Multi-Component Tree-Adjoining Grammars
with Tree Tuples \cite{Lichte:07}). More precisely, tree-based
grammars are first converted into equivalent RCGs, which are then used
for parsing. The result of RCG parsing is finally interpreted to extract
a derivation structure for the input grammar, as well as to perform
additional processings (e.g., semantic calculus, extraction of
dependency views).

The paper is structured as follows. In section~\ref{sec:rcg}, we
present the architecture of the TuLiPA parsing environment and show
how the use of RCG as a pivot formalism makes it easier to design a
modular system that can be extended to support several dimensions
(syntax, semantics) and/or formalisms. In
section~\ref{sec:environment}, we give some desiderata for grammar
engineering and present TuLiPA's current state with respect to 
these. In section~\ref{sec:comparison}, we compare this system with
existing approaches for parsing and more generally for grammar
engineering. Finally, in section~\ref{sec:future}, we conclude by
presenting future work.

\section{Range Concatenation Grammar as a pivot formalism}
\label{sec:rcg}
The main idea underlying TuLiPA is to use RCG as a pivot formalism for
RCG has appealing formal properties (e.g., a generative capacity
lying beyond Linear Context Free Rewriting Systems and a polynomial
parsing complexity) and there exist efficient algorithms, for RCG
parsing \cite{Boullier:00} and for grammar transformation into RCG
\cite{Boullier:98,Boullier:99}.

Parsing with TuLiPA is thus a 3-step process:
\begin{enumerate}
\item The input tree-based grammar is converted into an RCG (using the
algorithm of Kallmeyer and Parmentier
\shortcite{KallmeyerParmentier:08} when dealing with TT-MCTAG). 
\item The resulting RCG is used for parsing the input string using an
extension of the parsing algorithm of Boullier \shortcite{Boullier:00}. 
\item The RCG derivation structure is interpreted to extract the
derivation and derived trees with respect to the input grammar.
\end{enumerate}

The use of RCG as a pivot formalism, and thus of an RCG parser as a
core component of the system, leads to a modular architecture. In
turns, this makes TuLiPA more easily extensible, either in terms of
functionalities, or in terms of formalisms. 

\subsection{Adding functionalities to the parsing environment}
As an illustration of TuLiPA's extensibility, one may consider two
extensions applied to the system recently.

\begin{figure}[ht]
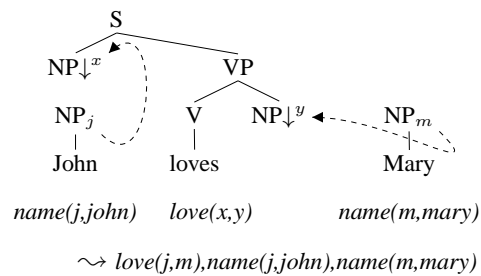

  \begin{center}
    {\small
      \begin{tabular}{ccccc}
        \multicolumn{2}{c}{\node{s}{S}}&\\[2ex]
        \node{np1}{NP$\downarrow^{x}$}
        &\multicolumn{2}{c}{\node{vp1}{VP}}
        \\[2ex]
        \node{np3}{NP$_{j}$}
        &\node{v1}{V}
        &\node{np2}{NP$\downarrow^{y}$} 
        &\node{np4}{NP$_{m}$}
        \\[2ex]
        \node{jon}{John} 
        &\node{love}{loves} 
        &&
        \node{mary}{Mary} 
        \\[2ex]
          {\small {\it name(j,john)}}
          &\multicolumn{2}{l}{{\small {\it love(x,y)}}}
          &{\small {\it name(m,mary)}}
      \end{tabular}
      \nodeconnect{s}{np1} \nodeconnect{s}{vp1}
      \nodeconnect{vp1}{v1} \nodeconnect{vp1}{np2}
      \nodeconnect{v1}{love} \nodeconnect{np3}{jon}
      \nodeconnect{np4}{mary}
    }
    {\dotted
      \anodecurve[br]{np3}[tr]{np1}{0.4in}
      \anodecurve[br]{np4}[r]{np2}{0.4in}
    }
    \vspace{0.3cm}
    
    $\leadsto$ {\small {\it love(j,m),name(j,john),name(m,mary)}}
  \end{center} 
  
  \caption{\label{fig:sem} Semantic calculus in Feature-Based TAG.}
\end{figure}

First, a semantic calculus using the syntax/semantics interface for
TAG proposed by Gardent and Kallmeyer \shortcite{GardentKallmeyer:03}
has been added. This interface associates each tree with flat semantic
formulas. The arguments of these formulas are unification variables,
which are co-indexed with features labelling the nodes of the
syntactic tree. During classical TAG derivation, trees are combined,
triggering unifications of the feature structures labelling nodes. As
a result of these unifications, the arguments of the semantic formulas
are unified (see Fig.~\ref{fig:sem}). 

In our system, the semantic support has been integrated by
(i)~extending the internal tree objects to include semantic 
formulas (the RCG-conversion is kept unchanged), and (ii)~extending
the construction of the derived tree (step 3) so that during the
interpretation of the RCG derivation in terms of tree combinations,
the semantic formulas are carried and updated with respect to the
feature unifications performed. 

\begin{figure}[ht]
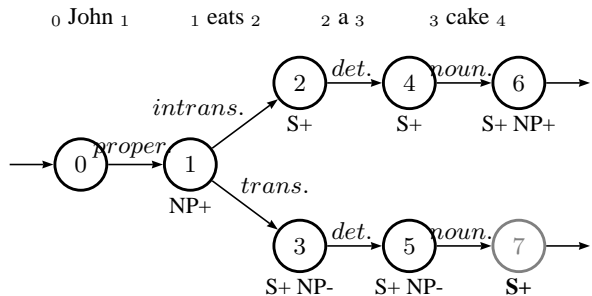

  \begin{center}
    
{\small
    \begin{tabular}{p{1.8cm}p{1.3cm}p{1cm}p{2cm}}
      ~~~~~$_0$~John~$_1$ & $_1$~eats~$_2$ & $_2$~a~$_3$ & $_3$~cake~$_4$ \\[2ex]
    \end{tabular}
      
    \VCDraw[.6]{
      \begin{VCPicture}{(-2,1)(18,8)}
        \LargeState\State[0]{(0,4)}{i} 
        \LargeState\State[1]{(4,4)}{j}
        \VCPut[0]{(4,2.5)}{{\Large NP+}}
        \LargeState\State[2]{(8,7)}{e1}
        \VCPut[0]{(8,5.5)}{{\Large S+}}
        \LargeState\State[3]{(8,1)}{e2}
        \VCPut[0]{(8,-0.5)}{{\Large S+~NP-}}
        \LargeState\State[4]{(12,7)}{d1}
        \VCPut[0]{(12,5.5)}{{\Large S+}}
        \LargeState\State[5]{(12,1)}{d2}
        \VCPut[0]{(12,-0.5)}{{\Large S+~NP-}}
        \LargeState\State[6]{(16,7)}{c1}
        \VCPut[0]{(16,5.5)}{{\Large S+~NP+}}
        \DimState\LargeState\State[7]{(16,1)}{c2}
        \VCPut[0]{(16,-0.5)}{{\Large {\bf S+}}}
        \Initial[w]{i}
        \Final{c1}
        \Final{c2}
        \EdgeL{i}{j}{proper.}
        \EdgeL{j}{e1}{intrans.}
        \EdgeL{j}{e2}{trans.}
        \EdgeL{e1}{d1}{det.}
        \EdgeL{e2}{d2}{det.}
        \EdgeL{d1}{c1}{noun.}
        \EdgeL{d2}{c2}{noun.}
      \end{VCPicture}
    }

}
  \end{center}
  
  \vspace{0.5cm}

  \caption{\label{fig:automaton} Polarity-based lexical disambiguation.}
\end{figure}

Secondly, let us consider lexical disambiguation. Because of the high
redundancy lying within lexicalized formalisms such as lexicalized
TAG, it is common to consider tree schemata having a frontier node
marked for {\it anchoring} (i.e., lexicalization). At parsing
time, the tree schemata are anchored according to the input
string. This anchoring selects a subgrammar supposed to cover the
input string. Unfortunately, this subgrammar may contain many trees
that either do not lead to a parse or for which we know {\it a priori}
that they cannot be combined within the same derivation (so we
should not predict a derivation from one of these trees to another
during parsing). As a result, the parser could have poor performance
because of the many derivation paths that have to be explored. 
Bonfante et al.~\shortcite{bonfante04polarization} proposed to
polarize the structures of the grammar, and to apply an automaton-based
filtering of the compatible structures. The idea is the following. One 
compute polarities representing the needs/resources brought by a
given tree (or tree tuple for TT-MCTAG). A substitution or foot node
with category NP reflects a need for an NP (written NP-). In the same
way, an NP root node reflects a resource of type NP (written
NP+). Then you build an automaton whose edges correspond to trees, and
states to polarities brought by trees along the path. The automaton is
then traversed to extract all paths leading to a final state with a
neutral polarity for each category and +1 for the axiom (see
Fig.~\ref{fig:automaton}, the  state 7 is the only valid state and
\{proper., trans., det., noun.\} the only compatible set of trees). 

In our context, this polarity filtering has been added before step 1,
leaving untouched the core RCG conversion and parsing steps. The idea
is to compute the sets of compatible trees (or tree tuples for
TT-MCTAG) and to convert these sets separately. Indeed the RCG has to
encode only valid adjunctions/substitutions. Thanks to this automaton-based
``clustering'' of the compatible tree (or tree tuples), we avoid
predicting incompatible derivations. Note that the time saved by using a
polarity-based filter is not negligible, especially when parsing long 
sentences.\footnote{An evaluation of the gain brought by this technique
when using Interaction Grammar is given by Bonfante et
al.~\shortcite{bonfante04polarization}.}

\subsection{Adding formalisms to the parsing environment}
Of course, the two extensions introduced in the previous section may
have been added to other modular architectures as well. The main gain
brought by RCG is the possibility to parse not only tree-based
grammars, but other formalisms provided they can be encoded into
RCG. In our system, only TAG and TT-MCTAG have been considered so
far. Nonetheless, Boullier \shortcite{Boullier:98} and S{\o}gaard
\shortcite{soegaard07phd} have defined transformations into RCG for
other mildly context-sensitive formalisms.\footnote{These include
  Multi-Component Tree-Adjoining Grammar, Linear Indexed Grammar, Head
  Grammar, Coupled Context Free Grammar, Right Linear Unification
  Grammar and Synchronous Unification Grammar.}

\begin{figure}[ht]
  \begin{center}
    \includegraphics[width=0.48\textwidth]{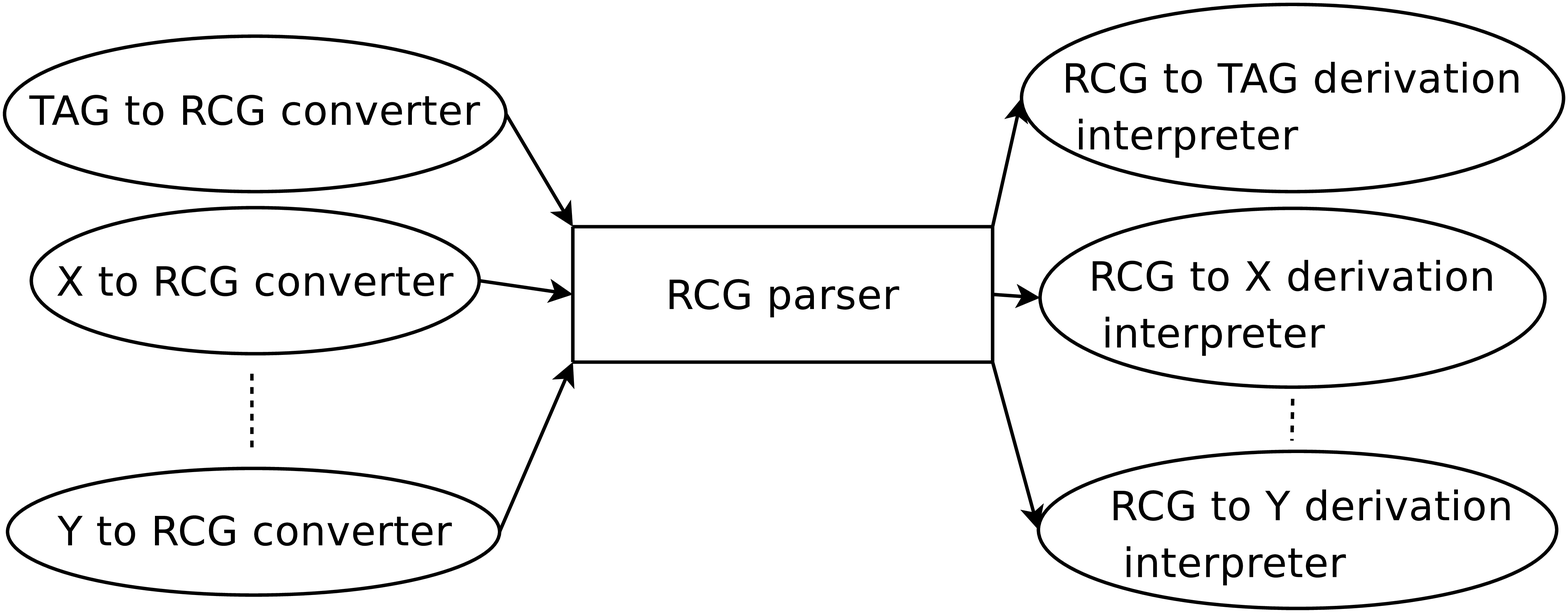}
  \end{center}
  \caption{\label{fig:Xtulipa} Towards a multi-formalism parsing environment.}
\end{figure}

To sum up, the idea would be to keep the core RCG parser, and to
extend TuLiPA with a specific conversion module for each targeted
formalism. On top of these conversion modules, one should also provide
interpretation modules allowing to decode the RCG derivation forest in
terms of the input formalism (see Fig.~\ref{fig:Xtulipa}).

An important point remains to be discussed. It concerns the role of
lexicalization with respect to the formalism used. Indeed, the
tree-based grammar formalisms currently supported (TAG and TT-MCTAG)
both share the same lexicalization process (i.e., tree
{\it anchoring}). Thus the lexicon format is common to these formalisms. As
we will see below, it corresponds to a 2-layer lexicon made of
inflected forms and lemma respectively, the latter selecting specific
grammatical structures. When parsing other formalisms, it is still
unclear whether one can use the same lexicon format, and if not what
kind of general lexicon management module should be added to the 
parser (in particular to deal with morphology).

\section{Towards a complete grammar engineering environment}
\label{sec:environment}
So far, we have seen how to use a generic parsing architecture relying
on RCG to parse different formalisms. In this section, we adopt a
broader view and enumerate some requirements for a linguistic
resource development environment. We also see to what extent these
requirements are fulfilled (or partially fulfilled) within the TuLiPA
system.

\subsection{Grammar engineering with TuLiPA}
As advocated by Erbach \shortcite{erbach92tools}, grammar engineering needs
{\it ``tools for testing the grammar with respect to consistency, coverage,
overgeneration and accuracy''}. These characteristics may be taken into
account by different interacting software. Thus, consistency can be
checked by a semi-automatic grammar production device, such as the XMG
system of Duchier et al.~\shortcite{dlp04}. Overgeneration is
mainly checked by a generator (or by a parser with adequate test
suites), and coverage and accuracy by a parser. In our case, the TuLiPA
system provides an entry point for using a grammar production system
(and a lexicon conversion tool introduced below), while including a
parser. Note that TuLiPA does not include any generator, nonetheless
it uses the same lexicon format as the GenI surface realizer for
TAG\footnote{\url{http://trac.loria.fr/~geni}}. 

TuLiPA's input grammar is designed using XMG, which is a {\it metagrammar}
compiler for tree-based formalisms. In other terms, the linguist
defines a factorized description of the grammar (the so-called
metagrammar) in the XMG language. Briefly, an XMG metagrammar consists 
of (i)~elementary tree fragments represented as tree description logic
formulas, and (ii)~conjunctive and disjunctive combinations of these
tree fragments to describe actual TAG tree schemata.\footnote{See 
  \cite{crabbe05grammatical} for a presentation on how to use the
  XMG formalism for describing a core TAG for French.}
This metagrammar is then compiled by the XMG system to produce a tree
grammar in an XML format. 
Note that the resulting grammar contains tree schemata (i.e.,
unlexicalized trees). To lexicalize these, the linguist defines a
lexicon mapping words with corresponding sets of trees. Following XTAG
\shortcite{xtag01lexicalized}, this lexicon is a 2-layer lexicon made
of morphological and lemma specifications. The motivation of this
2-layer format is (i)~to express linguistic generalizations at the
lexicon level, and (ii)~to allow the parser to only select a
subgrammar according to a given sentence, thus reducing parsing complexity.
TuLiPA comes with a lexicon conversion tool (namely lexConverter)
allowing to write a lexicon in a user-friendly text format and to
convert it into XML. An example of an entry of such a lexicon is
given in Fig.~\ref{fig:resources}. 

\begin{figure}[ht]
Morphological specification:\\
\begin{tabular}{|ccc|}
\hline
vergisst & vergessen & [pos=v,num=sg,per=3] \\
\hline 
\end{tabular} \\

Lemma specification:\\
\begin{tabular}{|l|}
\hline
$*$ENTRY: vergessen \\
$*$CAT: v \\
$*$SEM: BinaryRel[pred=vergessen] \\
$*$ACC: 1 \\
$*$FAM: Vnp2 \\
$*$FILTERS: [] \\
$*$EX: {} \\
$*$EQUATIONS: \\
NParg1 $\rightarrow$ cas = nom \\
NParg2 $\rightarrow$ cas = acc \\
$*$COANCHORS: \\
\hline
\end{tabular}

\caption{Morphological and lemma specification of {\it vergisst}.}
\label{fig:resources}
\end{figure}

The morphological specification consists of a word, the corresponding
lemma and morphological features.  
The main pieces of information contained in the lemma specification
are the $*$ENTRY field, which refers to the lemma, the $*$CAT field
referring to the syntactic category of the anchor node, the $*$SEM field
containing some semantic information allowing for semantic
instantiation, the $*$FAM field, which contains the name of the tree
family to be anchored, the $*$FILTERS field which consists of a feature
structure constraining by unification the trees of a given family that
can be anchored by the given lemma (used for instance for
non-passivable verbs), the $*$EQUATIONS field allowing for the definition
of equations targeting named nodes of the trees, and the $*$COANCHORS
field, which allows for the specification of co-anchors (such as 
{\it by} in the verb {\it to come by}).

From these XML resources, TuLiPA parses a string, corresponding either
to a sentence or a constituent (noun phrase, prepositional phrase,
{\it etc.}), and computes several output pieces of information, namely
(for TAG and TT-MCTAG): derivation/derived trees, semantic
representations (computed from underspecified representations using the utool
software\footnote{See~\url{http://www.coli.uni-saarland.de/projects/chorus/utool/},
  with courtesy of Alexander Koller.}, or dependency views of the derivation
trees (using the DTool software\footnote{With courtesy of Marco Kuhlmann.}). 

\subsection{Grammar debugging}

The engineering process introduced in the preceding section belongs to
a development cycle, where one first designs a grammar and corresponding
lexicons using XMG, then checks these with the parser, fixes them,
parses again, and so on.

To facilitate grammar debugging, TuLiPA includes both a verbose and a
robust mode allowing respectively to (i)~produce a log of the
RCG-conversion, RCG-parsing and RCG-derivation interpretation, and
(ii)~display mismatching features leading to incomplete derivations. 
More precisely, in robust mode, the parser displays derivations step
by step, highlighting feature unification failures.

\begin{figure}[ht]
  \begin{center}
    \includegraphics[width=0.48\textwidth]{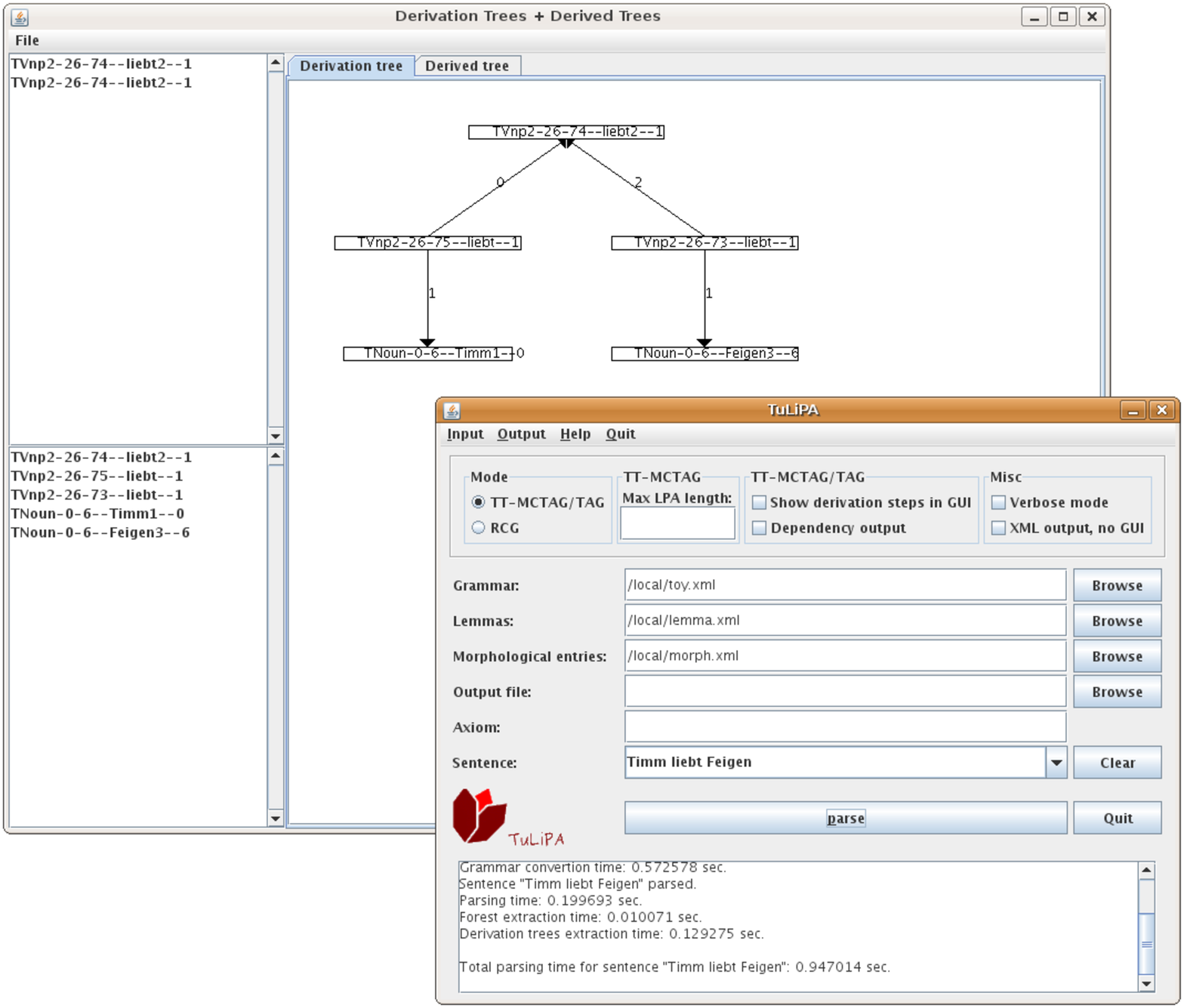}
  \end{center}
  \caption{\label{fig:gui} TuLiPA's Graphical User Interface.}
\end{figure}

TuLiPA's options can be activated via an intuitive Graphical
User Interface (see Fig.~\ref{fig:gui}).

\subsection{Towards a functional common interface}

\begin{figure}[ht]
  \begin{center}
    \includegraphics[width=0.48\textwidth]{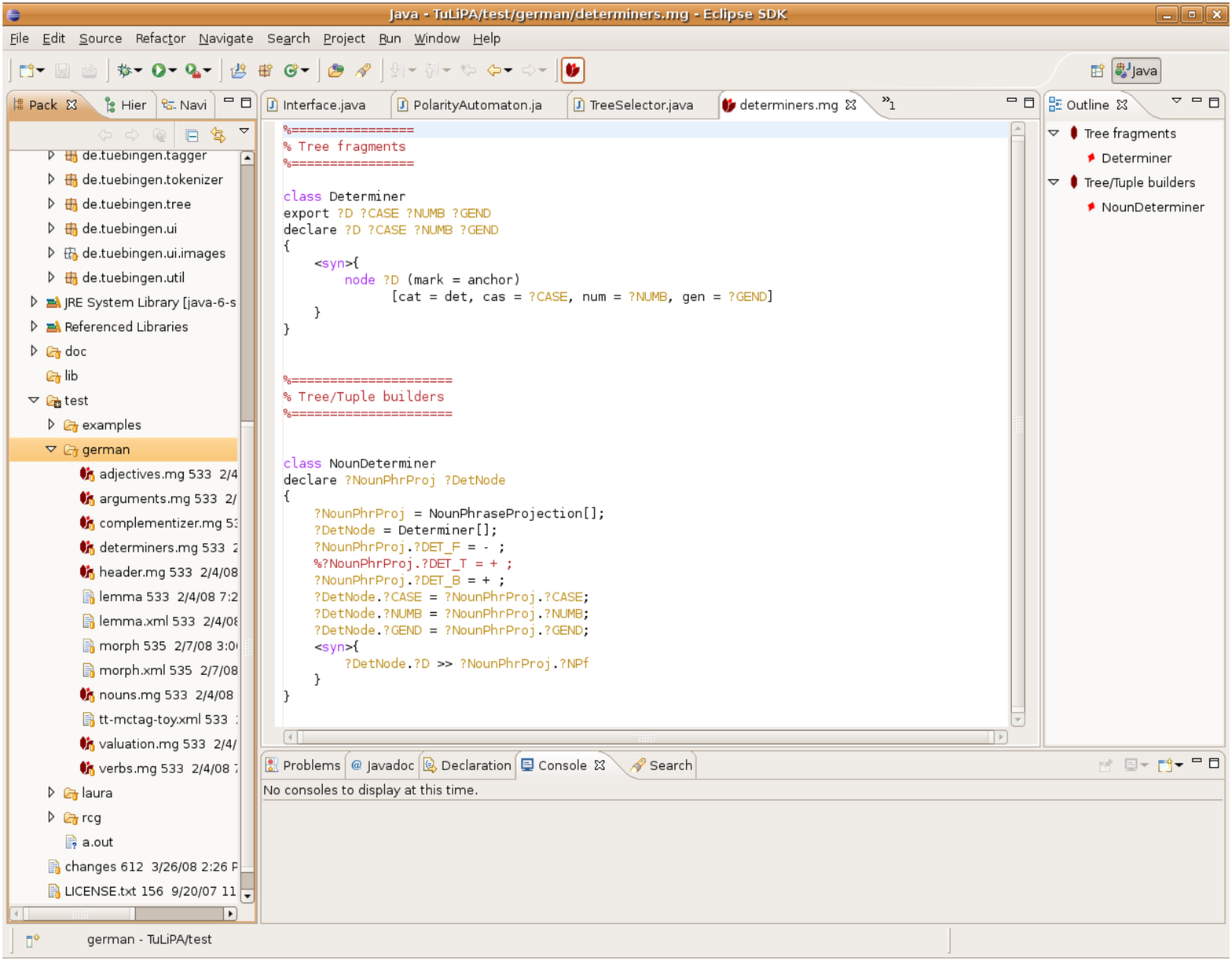}
  \end{center}
  \caption{\label{fig:eclipse} TuLiPA's eclipse plug-in.}
\end{figure}

Unfortunately, as mentioned above, the linguist has to move
back-and-forth from the grammar/lexicon descriptions to the parser,
i.e., each time the parser reports grammar errors, the linguist
fixes these and then recomputes the XML files and then parses again. 
To avoid this tedious task of resources re-compilation, we started
developing an Eclipse\footnote{See~\url{http://www.eclipse.org}}
plug-in for the TuLiPA system. Thus, the linguist will be able to
manage all these resources, and to call the parser, the metagrammar
compiler, and the lexConverter from a common interface (see
Fig.~\ref{fig:eclipse}). 

The motivation for this plug-in comes from the observation that
designing electronic grammars is a task comparable to designing source
code. A powerful grammar engineering environment should thus come with
development facilities such as precise debugging information, syntax
highlighting, {\it etc}. Using the Eclipse open-source development
platform allows for reusing several components inherited from the
software development community, such as plug-ins for version control,
editors coupled with explorers, {\it etc}.

Eventually, one point worth considering in the context of grammar
development concerns data encoding. To our knowledge, only few
environments provide support for UTF-8 encoding, thus guarantying the
coverage of a wide set of charsets and languages. In TuLiPA, we added
an UTF-8 support (in the lexConverter), thus
allowing to design a TAG for Korean (work in progress).

\subsection{Usability of the TuLiPA system}

As mentioned above, the TuLiPA system is made of several interacting
components, that one currently has to install separately. Nonetheless,
much attention has been paid to make this installation process as easy
as possible and compatible with all major platforms.\footnote{See
  \url{http://sourcesup.cru.fr/tulipa}.}

XMG and lexConverter can be installed by compiling their sources
(using a {\it make} command). TuLiPA is developed in Java and
released as an executable jar. No compilation is needed for it, the
only requirement is the Gecode/GecodeJ library\footnote{See
  \url{http://www.gecode.org/gecodej}.} (available as a binary package for
many platforms). Finally, the TuLiPA eclipse plug-in can be installed
easily from eclipse itself. All these tools are released under Free
software licenses (either GNU GPL or Eclipse Public License).

This environment is being used (i)~at the University of T\"ubingen, in
the context of the development of a TT-MCTAG for German describing both
syntax and semantics, and (ii)~at LORIA Nancy, in the development of an
XTAG-based metagrammar for English. The German grammar, called GerTT
(for German Tree Tuples), is released under a LGPL license for
Linguistic Resources\footnote{See
  \url{http://infolingu.univ-mlv.fr/DonneesLinguistiques/Lexiques-Grammaires/lgpllr.html}} 
and is presented in \cite{kallmeyer08developping}. The test-suite
currently used to check the grammar is hand-crafted. A more systematic
evaluation of the grammar is in preparation, using the 
Test Suite for Natural Language Processing \cite{lehmann96}.

\section{Comparison with existing approaches}
\label{sec:comparison}

\subsection{Engineering environments for tree-based grammar formalisms}

To our knowledge, there is currently no available parsing environment
for multi-component TAG.

Existing grammar engineering environments for TAG include the DyALog
system\footnote{See~\url{http://dyalog.gforge.inria.fr}} described in
Villemonte de la Clergerie \shortcite{DyALogCSLP05}. DyALog is a
compiler for a logic programming language using tabulation and dynamic
programming techniques. This compiler has been used to implement
efficient parsing algorithms for several formalisms, including TAG and
RCG. Unfortunately, it does not include any built-in GUI and requires
a good know\-led\-ge of the GNU build tools to compile parsers. This makes
it relatively difficult to use. DyALog's main quality lies in its
efficiency in terms of parsing time and its capacity to handle very
large resources. Unlike TuLiPA, it does not compute semantic
representations.

The closest approach to TuLiPA corresponds to the SemTAG 
system\footnote{See~\url{http://trac.loria.fr/~semconst}}, which
extends TAG parsers compiled with DyALog with a semantic calculus
module \cite{semtagACL07}. Unlike TuLiPA, this system only supports
TAG, and does not provide any graphical output allowing to easily
check the result of parsing.

Note that, for grammar designers mainly interested in TAG, SemTAG and
TuLiPA can be seen as complementary tools. Indeed, one may use TuLiPA
to develop the grammar and check specific syntactic structures thanks
to its intuitive parsing environment. Once the grammar is stable, one
may use SemTAG in batch processing to parse corpuses and build
semantic representations using large grammars. This combination of
these 2 systems is made easier by the fact that both use the same
input formats (a metagrammar in the XMG language and a text-based
lexicon). This approach is the one being adopted for the development
of a French TAG equipped with semantics.

For Interaction Grammar \cite{perrier00interaction}, there exists an
engineering environment gathering the XMG metagrammar compiler and an
eLEtrOstatic PARser
(LEOPAR).\footnote{See~\url{http://www.loria.fr/equipes/calligramme/leopar/}}
This environment is being used to develop an Interaction Grammar for
French. TuLiPA's lexical disambiguation module reuses techniques
introduced by LEOPAR. Unlike TuLiPA, LEOPAR does not currently support
semantic information.

\subsection{Engineering environments for other grammar formalisms}

For other formalisms, there exist state-of-the-art grammar engineering
environments that have been used for many years to design large
deep grammars for several languages. 

For Lexical Functional Grammar, one may cite the Xerox Linguistic
Environment
(XLE).\footnote{See~\url{http://www2.parc.com/isl/groups/nltt/xle/}}
For Head-driven Phrase Structure Grammar, the main available systems
are the Linguistic Knowledge Base 
(LKB)\footnote{See~\url{http://wiki.delph-in.net/moin}} and the TRALE
system.\footnote{See~\url{http://milca.sfs.uni-tuebingen.de/A4/Course/trale/}} 
For Combinatory Categorial Grammar, one may cite the OpenCCG
library\footnote{See~\url{http://openccg.sourceforge.net/}} and the
C\&C parser.\footnote{See~\url{http://svn.ask.it.usyd.edu.au/trac/candc/wiki}}

These environments have been used to develop broad-coverage resources
equipped with semantics and include both a generator and a
parser. Unlike TuLiPA, they represent advanced projects, that have
been used for dialog and machine translation
applications. They are mainly tailored for a specific
formalism.\footnote{Nonetheless,
  Beavers~\shortcite{beavers-documentation} encoded a CCG in the LKB's
  Type Description Language.}

\section{Future work}
\label{sec:future}

In this section, we give some prospective views concerning engineering
environments in general, and TuLiPA in particular. We first distinguish
between 2 main usages of grammar engineering environments, namely
a pedagogical usage and an application-oriented usage, and finally
give some comments about multi-formalism.

\subsection{Pedagogical usage}
Developing grammars in a pedagogical context needs facilities
allowing for inspection of the structures of the grammar, step-by-step
parsing (or generation), along with an intuitive interface. The idea is
to abstract away from technical aspects related to implementation
(intermediate data structures, optimizations, etc.).

The question whether to provide graphical or text-based editors can be
discussed. As advocated by Baldridge et
al.~\shortcite{Baldridge-et-al-GEAF07}, a low-level text-based 
specification can offer more flexibility and bring less frustration to
the grammar designer, especially when such a specification can be
graphically interpreted. This is the approach chosen by XMG, where the
grammar is defined via an (advanced or not) editor such as gedit or
emacs. Within TuLiPA, we chose to go further by using the Eclipse
platform. Currently, it allows for displaying a summary of the content
of a metagrammar or lexicon on a side panel, while editing these on a
middle panel. These two panels are linked via a jump functionality. 
The next steps concern (i)~the plugging of a graphical viewer to
display the (meta)grammar structures independently from a given parse,
and (ii)~the extension of the eclipse plug-in so that one can easily
consistently modify entries of the metagrammar or lexicon (especially
when these are split over several files).

\subsection{Application-oriented usage}
When dealing with applications, one may demand more from the grammar
engineering environment, especially in terms of efficiency and
robustness (support for larger resources, partial parsing, etc.).

Efficiency needs optimizations in the parsing engine making it
possible to support grammars containing several thousands of
structures. One interesting question concerns the compilation of a
grammar either off-line or on-line. In DyALog's approach, the grammar
is compiled off-line into a logical automaton encoding all possible
derivations. This off-line compilation can take some minutes with a
TAG having 6000 trees, but the resulting parser can parse sentences
within a second.

In TuLiPA's approach, the grammar is compiled into an RCG
on-line. While giving satisfactory results on reduced
resources\footnote{For a TT-MCTAG counting about 300 sets of trees and
  an and-crafted lexicon made of about 300 of words, a 10-word sentence
  is parsed (and a semantic representation computed) within seconds.},
it may lead to troubles when scaling up. This is especially true for
TAG (the TT-MCTAG formalism is by definition a factorized formalism
compared with TAG). In the future, it would be useful to look for a
way to pre-compile a TAG into an RCG off-line, thus saving the
conversion time.

Another important feature of grammar engineering environments consists
of its debugging functionalities. Among these, one may cite unit and
integration testing. It would be useful to extend the TuLiPA system to
provide a module for generating test-suites for a given grammar. The
idea would be to record the coverage and analyses of a grammar at a
given time. Once the grammar is further developed, these snapshots
would allow for regression testing.

\subsection{About multi-formalism}
We already mentioned that TuLiPA was opening a way towards
multi-formalism by relying on an RCG core. It is worth noticing that
the XMG system was also designed to be further extensible. Indeed, a
metagrammar in XMG corresponds to the combination of elementary
structures. One may think of designing a library of such structures,
these would be dependent on the target grammar formalism. The
combinations may represent general linguistic concepts and would be
shared by different grammar implementations, following ideas presented
by Bender et al.~\shortcite{bender05shared}.

\section{Conclusion}
\label{sec:conclusion}
In this paper, we have presented a multi-formalism parsing architecture
using RCG as a pivot formalism to parse mildly context-sensitive
formalisms (currently TAG and TT-MCTAG). 
This system has been designed to facilitate grammar development by
providing user-friendly interfaces, along with several functionalities
(e.g.,~dependency extraction, derivation/derived tree display and
semantic calculus). It is currently used for developing a core
grammar for German.

At the moment, we are working on the extension of this architecture to
include a fully functional Eclipse plug-in. Other current tasks concern
optimizations to support large scale parsing and the extension of the
syntactic and semantic coverage of the German grammar under development. 

In a near future, we plan to evaluate the parser and the German grammar
(parsing time, correction of syntactic and semantic outputs) with
respect to a standard test-suite such as the TSNLP \cite{lehmann96}.

\section*{Acknowledgments}
This work has been supported by the Deutsche Forschungsgemeinschaft
(DFG) and the Deutscher Akademischer Austausch Dienst (DAAD, grant
A/06/71039).
We are grateful to three anonymous reviewers for valuable comments on
this work.

\bibliographystyle{coling}
\bibliography{abstract}

\end{document}